\documentclass{article}

\usepackage{microtype}
\usepackage{graphicx}
\usepackage{subfigure}
\usepackage{booktabs} 
\usepackage{multirow}
\usepackage{bm}

\usepackage{hyperref}

\usepackage[accepted]{icml2021}

\usepackage{amsmath}
\usepackage{amssymb}

\newtheorem{assumption}{Assumption}
\newtheorem{proposition}{Proposition}
\newtheorem{lemma}{Lemma}

\icmltitlerunning{Fine-Grained Gradient Restriction}

\begin{document}

\global\long\def\th{\boldsymbol{\theta}}%
\global\long\def\th{\boldsymbol{\theta}}%
\global\long\def\L{\mathcal{L}}%
\global\long\def\S{\boldsymbol{S}}%
\global\long\def\H{\text{H}}%
\global\long\def\R{\mathbb{R}}%
\global\long\def\Th{\boldsymbol{\Theta}}%
\global\long\def\RR{\mathfrak{R}}%
\global\long\def\E{\mathbb{E}}%
\global\long\def\Z{\mathcal{Z}}%
\global\long\def\T{\mathcal{T}}%
\global\long\def\D{\mathcal{D}}%
\global\long\def\X{\mathcal{X}}%
\global\long\def\Y{\mathcal{Y}}%
\global\long\def\P{\mathbb{P}}%
\global\long\def\z{\boldsymbol{z}}%
\global\long\def\g{\boldsymbol{g}}%
\global\long\def\G{\textbf{G}}%
\global\long\def\H{\textbf{H}}%
\global\long\def\v{\boldsymbol{v}}%
\global\long\def\q{\boldsymbol{q}}%

\twocolumn[
\icmltitle{
Fine-Grained Gradient Restriction: A Simple Approach\\ for Mitigating Catastrophic Forgetting
}

%



\icmlsetsymbol{equal}{*}

\begin{icmlauthorlist}
\icmlauthor{Bo Liu}{ut}
\icmlauthor{Mao Ye}{ut}
\icmlauthor{Peter Stone}{ut,sony}
\icmlauthor{Qiang Liu}{ut}
\end{icmlauthorlist}

\icmlaffiliation{ut}{The University of Texas at Austin}
\icmlaffiliation{sony}{Sony AI}

\icmlcorrespondingauthor{Bo Liu}{bliu0201@gmail.com}

\icmlkeywords{Machine Learning, ICML}

\vskip 0.3in
]



\printAffiliationsAndNotice{} 

\begin{abstract}
A fundamental challenge in continual learning is to balance the trade-off between learning new tasks and remembering the previously acquired knowledge. 
Gradient Episodic Memory (GEM)~\citep{lopez2017gradient} achieves this balance by utilizing a subset of past training samples to restrict the update direction of the model parameters.
In this work, we start by analyzing a often overlooked hyper-parameter in GEM, the memory strength, which boosts the empirical performance by further constraining the update direction. 
We show that memory strength is effective mainly because it improves GEM's generalization ability and therefore leads to a more favorable trade-off.
By this finding, we propose two approaches that more flexibly constrain the update direction. Our methods are able to achieve uniformly better Pareto Frontiers of remembering old and learning new knowledge than using the memory strength. We further propose a computationally efficient method to approximately solve the optimization problem with more constraints.

\end{abstract}

\section{Introduction}
Intelligent creatures can learn continually over their lifetime, 
which helps them survive in the constantly changing world.
Continual learning (CL) studies the very problem of learning 
continually over a sequence of different tasks with a limited memory.
The main challenge of CL is to strike a balance between   remembering the previously acquired knowledge and learning new knowledge for new tasks.
Given a parameterized model, e.g. a deep neural network, at one extreme, directly learning on a new task with gradient descent will often drastically degrade the model's performance on previously learned tasks.
This phenomenon is called \emph{catastrophic forgetting}.
At the other extreme, the model will obviously not forget the learned task if we fix its parameters, but as a consequence, it will not learn any future task.
This phenomenon is called \emph{intransigence}~\citep{chaudhry2018riemannian}, i.e., the inability to update knowledge for a new task. With a limited memory, CL algorithms applied on a fixed-size model must seek an optimal trade-off between conflicting forgetting and intransigence.
\begin{figure}
    \centering
    \includegraphics[width=\columnwidth]{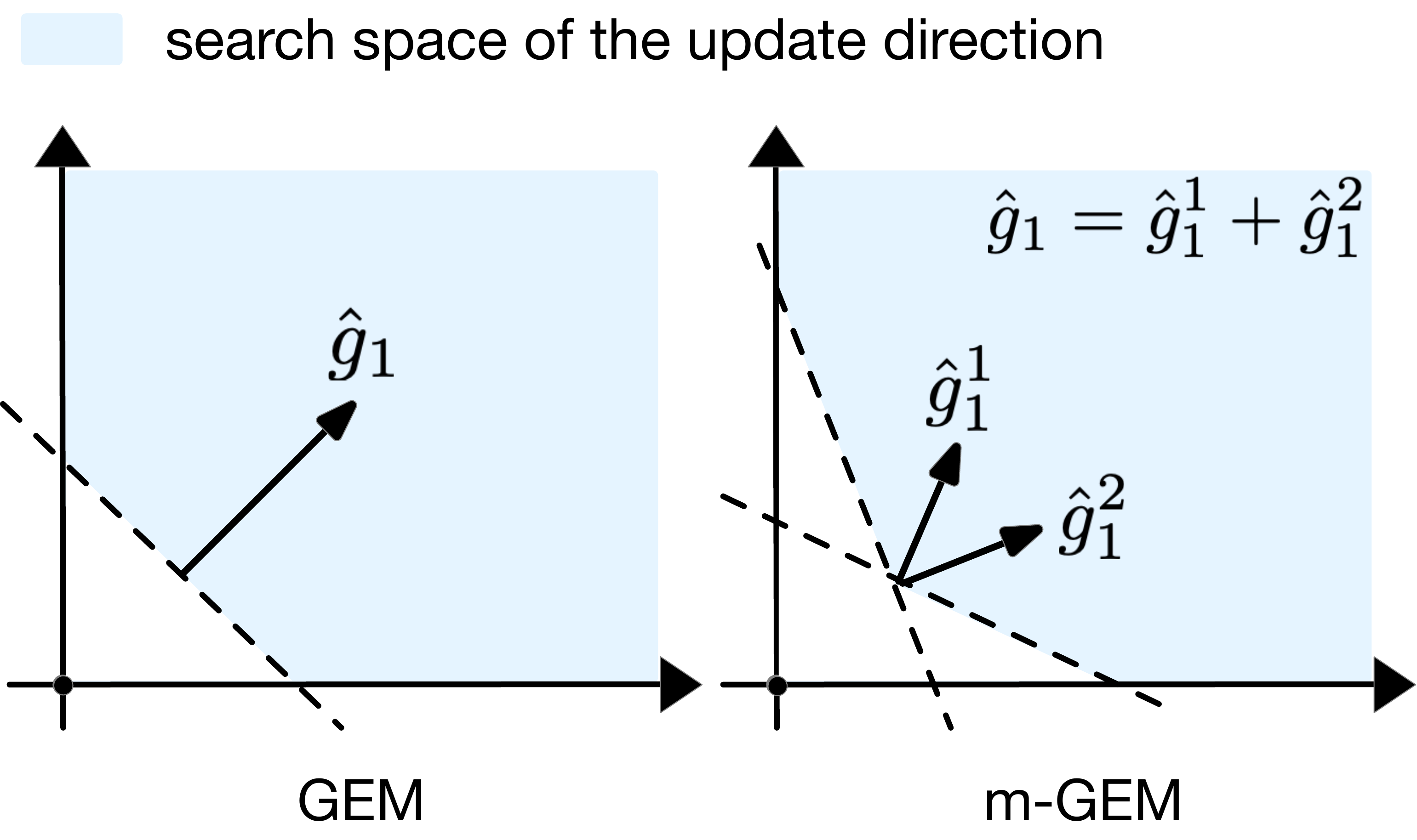}
    \vspace{-20pt}
    \caption{When progressing from task 1 to task 2, GEM computes $\tilde{g}_1$, the gradient on the episodic memory stored for task $1$. Then GEM ensures that the update direction is within the half-space defined by $\tilde{g}_1$. We propose to decompose $\tilde{g}_1$ into multiple vectors, i.e. $\tilde{g}_1^1, \tilde{g}_1^2$. The search space then becomes more constrained based on how we divide $\tilde{g}_1$.
    }
    \label{fig:intro}
    \vspace{-15pt}
\end{figure}

One popular method to trade-off forgetting and intransigence in CL is gradient episodic memory (GEM)~\citep{lopez2017gradient}. Intuitively, at each iteration of gradient descent, GEM searches an update direction in a space where its inner products with the gradients on past tasks are non-negative (See Figure~\ref{fig:intro}). Thus every direction in that search space does not increase the loss of past tasks. Within that search space, the update direction that minimizes its distance to the gradient of the currently learning task is used to update the model. As we are not able to store every observed data, the gradients on past tasks are approximated by the gradients calculated on sub-sampled data points from each of the prior tasks, which are called the episodic memories.

Ideally, with infinite size episodic memories, we can expect that GEM can learn without forgetting. However, in practice, as episodic memories are limited in size, GEM only ensures that loss computed on the episodic memory will not increase, which not necessarily implies the loss on past tasks will not increase. In other words, there exists a generalization gap due to the fact that episodic memories cannot cover the entire observed data.

In this work, we find that this generalization error can be effectively reduced by enforcing stronger and non-trivial constraints on the search space of the update direction. Specifically, we propose to divide the stored data points in the episodic memory into multiple splits and form new gradient constraints on each split, or separately apply GEM constraints to different parts of the underlying model parameters.
Our method is able to reduce catastrophic forgetting without exacerbating intransigence much and therefore achieves a better trade-off of them. To mitigate the overhead from introducing new constraints, we propose a computationally efficient method to approximately solve the optimization problem with more constraints. We call our algorithm modular GEM (mGEM). We demonstrate that mGEM can consistently achieve better Pareto Frontiers of remembering old and learning new knowledge than GEM, on both standard CL benchmarks and multi-domain datasets.
\section{Background}
In the continual learning problem, the learning machine will observe a sequence of data
\[
(x_{1},t_{1},y_{1}),...,(x_{i},t_{i},y_{i}),...,(x_{n},t_{n},y_{n}),
\]
where 
$x_i$ and $y_i$ are the input features and the corresponding
labels, and $t_{i}\in\mathbb{Z}$ is an integer task descriptor 
identifying the associated task.
For each task $t$, we assume the
input features and labels come from a joint distribution $(x,y)\sim\D_{t}$.
For different tasks $t$ and $t'$, $\D_t$ and $\D_{t'}$ are different in general.  
We adopt the standard assumption~\cite{lopez2017gradient} that every triplet $(x_{i},t_{i},y_{i})$ is \emph{locally} identically and independently distributed, i.e. $(x_{i},y_{i})\overset{\text{i.i.d.}}{\sim}\D_{t_{i}}$.
Our goal is to construct a model $f_{\theta}:\X\to\Y$ parameterized
by $\theta$ that is able to accurately predict $y$ given $x$, while $f$ can only go over the stream of data in an online fashion with a limited memory.

The key challenge of continual learning comes from the fact that examples are not drawn i.i.d. from a fixed probability distribution across different tasks.
In particular, the model loses its access to prior data once it starts to learn on a new task. The consequence of this non i.i.d.
pattern is often catastrophic forgetting, i.e. learning new tasks may significantly hurt the learner's performance on old tasks. To achieve better overall performance, the model needs to simultaneously prevent catastrophic forgetting and enable learning of new knowledge. Therefore, continual learning algorithms can be seen as attempting to improve the \emph{Pareto frontier} of these two conflicting objectives. In this paper, following \citep{riemer2018learning}, we measure the ability of learning new task by forward transfer and the ability of improving the performance of learned past task after learning a new task by backward transfer (see Eq~\ref{equ: transfer} for details).

\vspace{-5pt}
\paragraph{Notation}
We denote the integer set $\{1,2,....n\}$ as $[n]$. The $\ell_2$ vector norm is denoted as $\left\Vert \cdot\right\Vert$. Given a dataset $\D$, we define the loss over $\D$ by $\L_\D[f] = \E_{(x,y) \sim \D} \ell(f; x, y)$, where $\ell(f;x, y)$ is the loss of the learning machine $f$ on data pair $(x,y)$. For any two vectors $\boldsymbol{h}$, $\boldsymbol{h}'$ with the same dimension, we denote $\boldsymbol{h} \ge \boldsymbol{h}$ if $\boldsymbol{h}_i \ge \boldsymbol{h}'_i$ for every coordinate $i$. We denote the Rademacher complexity with sample size $n$ of set $S$ as $\RR_n[S]$. Given any set $S$, we denote its cardinality as $|S|$.

\subsection{Gradient of Episodic Memory}
In practice, the learning machine is usually trained with gradient descent. Given the loss $\L_\D[f_{\th}]$ on some dataset $\D$, suppose the learning machine is parameterized by $\th$, using the gradient descent, the parameter is updated iteratively by
\[
\th \leftarrow \th - \epsilon \g,
\]
where $\g=\nabla_{\th} \L_{\D}$ is the gradient of loss at current iteration and $\epsilon$ is the learning rate. Naively applying gradient descent for the continual learning problem is sub-optimal due to the catastrophic forgetting. To alleviate forgetting, Gradient Episodic Memory
(GEM) \cite{lopez2017gradient} searches a different updating direction $\z$ that does not increase the loss on prior tasks and decreases the loss on the current task as much as possible.

To achieve this, before learning a task $t$, GEM allocates an ``episodic memory" $\hat{\D}_{s}$, a subset of observed examples from $\D_s$, for every past task $s \in [t-1]$. At each iteration, GEM replaces $\g$ with a modified update direction $\z$ by solving the following constrained optimization problem
\begin{align}\label{opt: gem_primal}
 & \min_{\z}\left\Vert \g_{t}-\z\right\Vert ^{2} \ \ \text{s.t.\ } \left\langle \hat{\g}_{s},\z\right\rangle \ge0,\ \forall s<t.
\end{align}
Here we define $\g_{t}=\nabla_{\th}\L_{\D_{t}}[f_{\th}]$, the gradient computed for the current task $t$. Moreover, $\hat{\g}_{s}=\nabla_{\th}\L_{\hat{\D}_{s}}[f_{\th}]$ is the gradient computed on the episodic memory $\hat{\D}_s$ from the past task $s$. Finally, the model is updated with the solution $\z^*$ of (\ref{opt: gem_primal}) in place of the vanilla gradient
\[
\th \leftarrow\th-\epsilon\z.
\]
To see the intuition behind Equation~(\ref{opt: gem_primal}), with the small learning rate $\epsilon$, using the Taylor expansion, we obtain that for any prior task $s$,
\begin{equation}
\label{eq:taylor1}
\L_{\hat{\D}_{s}}\left[f_{\th}\right] - \L_{\hat{\D}_{s}}\left[f_{\th - \epsilon\z}\right] \approx \langle \hat{\g}_s,\z \rangle + \mathcal{O}\left(\epsilon^{2}\right),
\end{equation}
and for the current task $t$,
\begin{equation}
\label{eq:taylor2}
    \L_{\D_{t}}\left[f_{\th}\right] - \L_{\D_{t}}\left[f_{\th - \epsilon\z}\right] \approx  \langle \g_t,\z \rangle + \mathcal{O}\left(\epsilon^{2}\right).
\end{equation}
If we ignore the smaller order term $\mathcal{O}(\epsilon^2)$, constraints in Equation~(\ref{opt: gem_primal}) ensures that $\langle \hat{\g}_s,\z \rangle \geq 0$. Meanwhile, the objective in Equation~(\ref{opt: gem_primal}) ensures that $\langle \g_t,\z \rangle$ is large.

The decision variable $\z$ in the primal problem~(\ref{opt: gem_primal}) has the same size of the model's parameters, which can be millions for a deep neural network. Therefore, in practice, GEM solves the corresponding dual problem
\begin{align}\label{opt: gem_dual}
\min_{\v}\  & \frac{1}{2}\left\Vert \hat \G{}^{\top}\v+\g_t\right\Vert ^{2},\ \ \text{s.t.\ } \v\ge\boldsymbol{0}.
\end{align}
Here $\hat{\G} = -[\hat{\g}_1,...,\hat{\g}_{t-1}]^\top$. Now the dual decision variable $\v$ only has $t-1$ dimensions, and (\ref{opt: gem_dual}) can be efficiently solved by standard quadratic programming solvers. Based on the strong duality, the solution to the primal problem is given by $\hat \G{}^{\top}\v^* + \g_t$ where $\v^*$ is the solution of~(\ref{opt: gem_dual}).

\paragraph{Memory Strength}
In actual implementation, GEM introduces an extra ``memory strength'' hyper-parameter, 
which yields an optimization with  a stronger constraint: 
\begin{align}\label{opt: gem_dual_ms}
\min_{\v}\  & \frac{1}{2}\left\Vert \hat{\G}_{t}^{\top}\v+\g_t\right\Vert ^{2},\ \ \text{s.t.\ } \v\ge \q,
\end{align} 
where $\q\ge\bm{0}$ is called the memory strength. Although it was only treated as an implementation detail in the paper, we find that the memory strength can boost up the performance of GEM. For example, for common benchmarks in CL (Section~\ref{sec:cl-benchmark}), memory strength improves the average accuracy after learning on all tasks by $\bm{1.85}$ (MNIST Permutations) and $\bm{1.55}$ (MNIST Rotations).

\subsection{The Pareto Frontier of Forward and Backward Transfer}
Our analysis in Section \ref{sec: theory} shows that the primal problem of (\ref{opt: gem_dual_ms}) is actually equivalent to 
\begin{align} \label{opt: gem_bias}
\min_{z\in\Z}\left\Vert \g_{t}-\z\right\Vert ^{2}\ \text{s.t.}\ \left\langle \hat{\g}_{s},\z\right\rangle \ge\gamma_s,\ \forall s\in[t-1],
\end{align}
where $\gamma_s$ are some positive constants depending on $q$. GEM with larger $q$ puts a stronger constraint on $\left\langle \hat{\g}_{s},\z\right\rangle$, which enforces its value to be larger and thus increases the backward transfer, e.g. lower loss on past tasks. In Section~\ref{sec: theory}, we show that this stronger constraint is important because it mitigates the generalization gap caused by using a subset of the whole data points as the episodic memory. On the other hand, with stronger constraints, the feasible set becomes smaller and thus returns a larger solution to the objective, e.g. loss on the new task is reduced less. Figure \ref{fig: pareto} (blue line) provides an example of such trade-off. Each line is obtained by varying the memory strength $\q$.

\section{Modular GEM}
GEM is able to achieve a good Pareto frontier between backward and forward transfer with the introduced memory strength. Motivated by this, we propose two new simple but effective ways to more flexibly constrain the search space of the update direction, which are shown to achieve an (uniformly) better Pareto frontier compared with GEM with memory strength (see Figure \ref{fig: pareto}).
GEM computes the gradient of the full model on the entire episodic memory, which results in one constraint per task. Our method splits that gradient into several gradients computed either from different parts of the model or from different divides of the episodic memory. Since essentially we subdivides the gradient into multiple modules, we name our method modular-GEM (mGEM).

\vspace{-5pt}
\paragraph{Parameter-wise mGEM}
Our first scheme is to partition the parameter $\th$ of neural network
into $D$ sub-coordinates: $\th=[\th^{1},\th^{2},...,\th^{D}]$.
Then we separately apply GEM to each sub-coordinate. We call this method p-mGEM.
\begin{equation}
\begin{split}
\label{opt:pmGEM}
 & \min_{\z}\left\Vert \g_{t}-\z\right\Vert ^{2}\\
\text{s.t.\ } & \left\langle \hat{\g}_{s}^{d},\z\right\rangle \ge \gamma_s^d,\ \forall s\in[t-1],\ d\in[D],
\end{split}
\end{equation}
where $\hat{\g}_{s}^{d}=\nabla_{\th^{d}}\L_{\hat{\D}_{s}}[f_{\th}]$. Notice that all $\hat{\g}_s^d$ can be computed by back-propagating \emph{once} on the full model.

\vspace{-5pt}
\paragraph{Data-wise mGEM}
Similarly, we can also partition the episodic memory into several
groups: $\hat{\D}_{s}=\cup_{d=1}^{D}\tilde{\D}_{s}^{d}$ and then separately apply GEM.
We call this method d-mGEM and consider the following problem
\begin{equation}
\label{opt:dmGEM}
\begin{split}
 & \min_{\z}\left\Vert \g_{t} - \z\right\Vert ^{2}\\
\text{s.t.\ } & \left\langle \tilde{\g}_{s}^{d},\z\right\rangle \ge \gamma_s^d,\ \forall s\in[t-1],\ d\in[D],
\end{split}
\end{equation}
where $\tilde{\g}_{s}^{d}=\nabla_{\th}\L_{\tilde{\D}_{s}^{d}}[f_{\th}]$. For data-wise mGEM, the computation time becomes $D$ times longer since we need to separately computes the gradient for each split of the episodic memory.

To summarize, both parameter-wise and data-wise mGEM result in more restricted search space than vanilla GEM when $D > 1$, and reduce to GEM when $D=1$. By adjusting the value of $\gamma_s^d$ (or the $\q$ in the dual problem), mGEM can also incorporate the idea of memory strength.

\subsection{mGEM in Practice}
\label{sec:approx-GEM}
Although introducing more constraints naturally leads to more restricted search space and thus helps prevent forgetting, it also introduces computation overhead. Therefore, we propose to approximately solve the optimization problems in (\ref{opt:pmGEM}) and (\ref{opt:dmGEM}). For each $d$, the optimization corresponding to $d$ mimics that of the original GEM method. Therefore we similarly solve the dual problem, which is of the form
\begin{align}\label{opt: mgem_dual}
\min_{\v}\  & \frac{1}{2}\left\Vert \hat{\G}{}^{\top} \v+ \g \right\Vert ^{2},\ \ \text{s.t.\ } \v\ge \q,~~\text{where}~\q>0.
\end{align}
Instead of solving the optimization exactly with a quadratic programming solver, we propose to solve the optimization approximately in two stages. We first ignore the constraints and solve the optimization in closed-form, then project the solution back to the feasible space. To be specific,
\begin{itemize}
    \item In stage one, we solve 
        \begin{align*}
            \min_{\bm{\nu}}\  & \frac{1}{2}\left\Vert \hat{\G}{}^{\top} \bm{\nu} + \g \right\Vert ^{2}.
        \end{align*}
        The optimal solution is $\bm{\nu}^* = -(\hat{\G}\hat{\G}{}^{\top})^{-1}\hat{\G}\g$. Recall that for task $t$, $\hat{\G} = - [\hat{\g}_1, \dots \hat{\g}_{t-1}]^{\top} \in \mathbb{R}^{(t-1)\times n}$, where $n$ denotes the parameter size of the model $f_{\th}$. Here computing the term $(\hat{\G}\hat{\G}{}^{\top})^{-1}$ takes roughly $\mathcal{O}(t^2n)$ time. As random high dimensional vectors have a near zero inner product, we simplify the calculation by approximating $(\hat{\G}\hat{\G}{}^{\top})^{-1}$ with its diagonal matrix, which results in the solution
        \begin{equation*}
            \begin{split}
            \tilde{\bm{\nu}}^* &= - \H\hat{\G}\g,~\text{where}\\
            \H &=
            \begin{bmatrix}
               (\hat{\g}_1^{\top}\hat{\g}_1)^{-1} & & \\
                & \ddots & \\
                & & (\hat{\g}_{(t-1)}^{\top}\hat{\g}_{(t-1)})^{-1}
            \end{bmatrix}.
            \end{split}
        \end{equation*}
        Note that $\H$ can be efficiently computed in a batch using standard deep learning libraries.
    \item In stage two, we project the solution back to the feasible space and ensure that $\v \geq \q$:
    \begin{equation*}
        \tilde{\v}^* = [{\v}^*_1, ..., {\v}^*_{t-1}]^\top,\ \ \text{where}\ {\v}^*_i = \max(\tilde{\bm{\nu}}^*_i, {\q}_i).
    \end{equation*}
\end{itemize}
To summarize, the final update direction is given by
\begin{equation}
    \label{eq:approx-GEM}
    \z = \hat{\G}^{\top}\max(-\H\hat{\G}\g,~\q) + \g_t.
\end{equation}

We then apply Equation~(\ref{eq:approx-GEM}) to all modules considered in mGEM.

\section{Analysis} \label{sec: theory}
In this section, we provide analysis on why the proposed mGEM is able to improve the performance. We divide this question into three steps. For simplicity, we consider the case that there is only two tasks: a past task $s$ that the machine has learned, and the current task $t$ that the machine is learning. Generalizing our analysis to an arbitrary finite number of tasks is straightforward.

\subsection{The generalization of GEM}
Despite the fact that GEM is guaranteed to prevent $\L_{\hat{\D}_{s}}$ from increasing, GEM is not able to ensure that $\L_{\D_{s}}$
is not increasing. The reason is that the the episodic memory is
only a subset of observed examples and thus there is a generalization
gap between the loss on $\hat{\D}_{s}$ and that on $\D_{s}$. Suppose
that we have $\left\langle \hat{\g}_{s},\z\right\rangle \ge\gamma$,
the uniform concentration inequality shows that given any $\delta>0$,
we have, with probability $1-\delta$, $\left\langle \g_{s},\z\right\rangle \ge\gamma-\Delta$,
where $\Delta$ is some bias induced by the complexity of the search space
of $\z$ and $\delta$. 
Specifically, we have the following result. 

\begin{assumption}
For all previous tasks, the true gradient norm on those tasks are bounded:
\begin{align*}
\left\Vert \g_s\right\Vert _{2,\infty}:=\sup_{(x,y)\in\D_{s},\th\in\boldsymbol{\Theta}}\left\Vert \nabla_{\th}\ell(f_{\th};x,y)\right\Vert <\infty.
\end{align*}
\end{assumption}
\begin{proposition}\label{prop: generalization}
Given some search space $\Z$ of $\z$, suppose $\z^{*}$ is the solution of the following problem 
\[
\min_{\z\in\Z}\left\Vert \g_{t}-\z\right\Vert ^{2}\ \text{s.t.\ }\left\langle \hat{\g}_{s},\z\right\rangle \ge0,\ \forall s\in[t-1].
\]
Given any $\delta>0$, with probability at least $1-\delta$, we have
\begin{align*}
&\left\langle \g_{s},\z^{*}\right\rangle   \ge\left\langle \hat{\g}_{s},\z^{*}\right\rangle - \Delta, & \\
& \Delta =
 2\RR_{\left|\hat{\D}_{s}\right|}\left[\Z\right]+\sup_{\z\in\Z}\left\Vert \z\right\Vert \left\Vert \g_{s}\right\Vert _{2,\infty}\sqrt{\log(1/\delta)/\left|\hat{\D}_{s}\right|},&
\end{align*}
where $\Delta$ is the generalization gap 
and $\RR_{\left|\hat{\D}_{s}\right|}\left[\Z\right]$ denotes the Rademacher complexity of set $\mathcal Z$.
\end{proposition}
The above Proposition characterizes the worst-case generalization property on the backward transferring. We can see that, to improve the backward transferring, we can increase $\left\langle \hat{\g}_{s},\z^{*}\right\rangle$, reduce the generalization gap, or do both.

\subsection{Understanding Memory Strength}
Our analysis on the generalization property explains why memory strength is an effective trick in GEM. We start with one of the most natural ways to improve the backward transferring: increasing $\left\langle \hat{\g}_{s},\z^{*}\right\rangle$. By applying a stronger constraint, we consider the problem (\ref{opt: gem_bias}) mentioned earlier.
Denote $\boldsymbol{\gamma}=[\gamma_1,...,\gamma_{t-1}]^\top$ is the given non-negative hyper-parameter. By encouraging $\left\langle \hat{\g}_{s},\z\right\rangle $ to be strictly positive, we are able to improve the probability of non-negative backward transfer according to Proposition \ref{prop: generalization}. The dual problem of (\ref{opt: gem_bias}) is 
\begin{align} \label{opt: gem_dual_bias}
 & \min_{\v}\left\Vert \hat{\G}_{t}^{\top}\v+\g_{t}\right\Vert^2 -\boldsymbol{\gamma}^{\top}\v,\ \ \text{s.t.}\ \v\ge\boldsymbol{0}.
\end{align}
Compared with the dual problem (\ref{opt: gem_dual_ms}) of GEM with memory strength, (\ref{opt: gem_dual_bias}) enforces a positive $\boldsymbol{\gamma}$ (memory strength) by adding a regularization term ($-\boldsymbol{\gamma}^\top\v$) in the objective. From this perspective, GEM with memory strength is equivalent to problem (\ref{opt: gem_bias}). For simplicity, in the rest of the paper, we also refer to (\ref{opt: gem_dual_bias}) as GEM with memory strength.

\subsection{Worst Case Analysis on Backward Transfer}
Following the analysis in Proposition \ref{prop: generalization}, we know that GEM reduces the generalization gap by encouraging a larger $\left\langle \hat{\g}_s,\z\right\rangle$. Similarly, mGEM also encourages a larger inner product and therefore better generalization. In particular, we have the following proposition
\vspace{-5pt}
\begin{proposition}
Suppose $\z^{*}$ is the solution of (\ref{opt: gem_primal}) and $\z_{\text{MS}}^{*}$
is the solution of (\ref{opt: gem_bias}) with memory strength $\gamma_{s}$. For parameter-wise GEM, suppose that $\sum_{d=1}^{D}\gamma_{s}^{d}\ge\gamma_{s}$
and returns solution $\z_{\text{pmGEM}}^{*}$ . For data-wise GEM,
suppose that $\min_{d\in[D]}\gamma_{s}^{d}\ge\gamma_{s}$, and returns
solution $\z_{\text{dmGEM}}^{*}$. Then we have 
\begin{align*}
\left\langle \hat{\g}_{s},\z_{\text{p-mGEM}}^{*}\right\rangle  & \ge\left\langle \hat{\g}_{s},\z_{\text{MS}}^{*}\right\rangle \ge\left\langle \hat{\g}_{s},\z_{\text{}}^{*}\right\rangle \\
\left\langle \hat{\g}_{s},\z_{\text{d-mGEM}}^{*}\right\rangle  & \ge\left\langle \hat{\g}_{s},\z_{\text{MS}}^{*}\right\rangle \ge\left\langle \hat{\g}_{s},\z_{\text{}}^{*}\right\rangle .
\end{align*}
Therefore, in terms of reducing the generalization gap, both p-mGEM and d-mGEM perform better than GEM with memory strength, which further is better than the vanilla GEM.
\end{proposition}
The above analysis provides a theoretical  insight on how mGEM and GEM with memory strength improve the backward transferring through a worse-case analysis.

\subsection{mGEM Achieves a Better Pareto Frontier}
\begin{figure}[t]
\begin{centering}
\includegraphics[width=0.9\columnwidth]{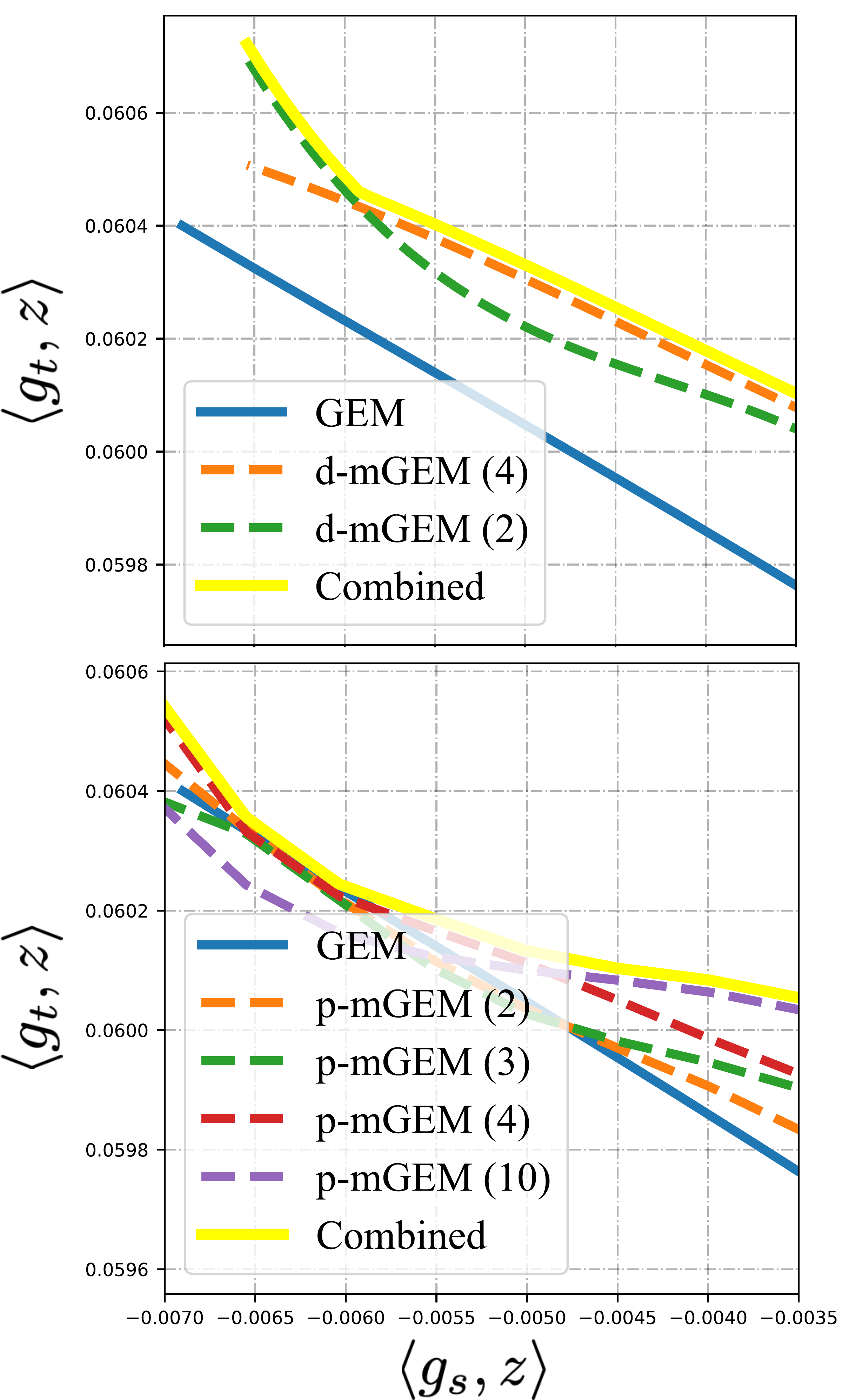}
\vspace{-5pt}
\caption{Trade-off between $\langle \g_s, \z \rangle $ (x-axis) vs. 
$\langle \g_t, \z \rangle $ (y-axis), where $\z$ is the update direction found by mGEM/GEM. Here, d/p-mGEM($n$) denotes there are $n$ modules in total. For example, p-mGEM(2) means we divide the model parameters into 2 groups and separately apply GEM on each.}
\label{fig: pareto}
\vspace{-10pt}
\end{centering}
\end{figure}

Compared with GEM with memory strength, mGEM gives two different ways to modify the constraints and thus can achieve a better Pareto frontier on forward and backward transfer. To show this, we apply p-mGEM and d-mGEM with different number of modules and different memory strengths to the Split Cifar100 benchmark. We plot the Pareto-frontier of the local backward and forward transfer. As described in Equation~(\ref{eq:taylor1}) and (\ref{eq:taylor2}), we measure the forward transfer by calculating the average value of $\left\langle \g_{t},\z\right\rangle$. To measure the backward transfer, we consider $\left\langle \g_{s},\z\right\rangle$, where the gradient on a past task is computed on the full training set rather than on episodic memories. We refer readers to the Appendix for more details. Figure \ref{fig: pareto} presents the result. We see that the backward transfer is larger when p-mGEM/d-mGEM have more modules and forward transfer is larger with fewer modules. Since the region corresponding to larger backward transfer (and thus less forgetting) is more important for typical CL problems, mGEM gives a clear improvement. 
Note that by allowing an adaptive number of modules, mGEM achieves a uniformly better Pareto frontier (yellow line in Figure \ref{fig: pareto}) than GEM.

\section{Related Work}
To prevent catastrophic forgetting in deep learning, three main categories of methods have been proposed recently.
\vspace{-8pt}
\paragraph{Regularization-based methods} Regularization-based methods introduce additional training objectives that constrain the model parameters to stay close to the previously learned ones. Popular methods often adopt a probabilistic perspective and view the old parameter as the prior on the model parameter before learning the new task. Specifically, EWC~\cite{kirkpatrick2017overcoming} uses the Fisher information matrix computed from previous tasks to regularize the learning. PI~\cite{zenke2017continual} computes a local "importance" measure for each neuron and the update is regularized according to that importance. RWALK~\cite{chaudhry2018riemannian} and VCL~\cite{nguyen2017variational} use the KL-divergence as the regularization. 
\vspace{-8pt}
\paragraph{Expandable architecture} Instead of regularizing the update direction for a fixed-size model, expandable architecture attempts to maximally re-use part of the parameters learned from previous tasks and introduces new parameters when intransigence happens. PROG-NN~\cite{rusu2016progressive} introduces a fixed number of neurons for each layer of the network and learns the lateral connections. DEN~\cite{yoon2017lifelong}, Learn-to-grow~\cite{li2019learn} and CPG~\cite{hung2019compacting} focus on more efficient growth of the network by allowing dynamic expansions.
\vspace{-8pt}
\paragraph{Memory-based approaches} A simple yet efficient way to prevent forgetting is to store a small subset of samples (i.e. the episodic memory) for memory consolidation. While methods like DGR~\cite{shin2017continual} learn a generative model that can generate artificial memories, they introduce the memory overhead of storing a generative model. On the other hand, GEM~\cite{lopez2017gradient} uses the episodic memory to setup a feasible space to search for new gradient update direction. OGD~\cite{farajtabar2020orthogonal} projects the gradient on the new task to a subspace where the projected gradient does not affect the model's output on old tasks. MER~\cite{riemer2018learning} is recently proposed to adopt a meta-learning objective to learn to not forget. As the optimal solution from GEM is a linear combination of gradients the current and previous tasks, MEGA~\cite{guo2019learning} studies how to form a better linear combination by considering the task losses. While MEGA also extends GEM, our method is orthogonal to MEGA. Besides studying the usage of episodic memories, \citet{aljundi2019gradient} also studies how to optimally select the them.

Our proposed method is closely related to GEM. However, we focus on explaining the trade-off of shrinking the feasible space in which the new update direction is searched. Particularly, besides how to pick and use the episodic memory, we focus on how these stored memories can form a better feasible space such that the found update direction achieves a better Pareto frontier.
\section{Experiments}
In this section, we first apply mGEM on the standard continual learning benchmarks and show that mGEM can outperform GEM. Then we further apply mGEM to datasets with more non-stationarity where each subtask is from a different domain. We compare mGEM against the following relevant baselines:
\vspace{-5pt}
\begin{itemize}
    \item \textbf{Single:} a single model that directly learns over the sequence of the task with stochastic gradient descent (SGD). It corresponds to one extreme where the model completely ignores forgetting and focuses only on learning the current task.
    \vspace{-2pt}
    \item \textbf{Independent:} an independent predictor for each task with fewer hidden neurons proportional to the number of tasks.
    \vspace{-2pt}
    \item  \textbf{EWC}: Elastic Weight Consolidation (EWC)~\cite{kirkpatrick2017overcoming} is a popular regularization-based method that adopts the Fisher Information matrix to prevent catastrophic forgetting. 
    \vspace{-2pt}
    \item \textbf{GEM}: Gradient Episodic Memory (GEM)~\cite{lopez2017gradient} is what we mainly compare against. GEM uses the episodic memory to compute a feasible space to search for the new update direction.
    \vspace{-2pt}
    \item \textbf{MER}: Meta Episodic Memory~\cite{riemer2018learning} a state-of-the-art episodic memory baseline that adopts meta learning objective to prevent forgetting.
\end{itemize}

Following~\citet{lopez2017gradient}, we report the final retained accuracy (ACC) and the backward transfer of knowledge (BWD) as the main metrics for comparison. Moreover, we also report the average learning accuracy, which is the average accuracy for each task right after it is learned, as a measure of the how much the model can learn forwardly on the new task (FWD). Specifically, assume there are $T$ tasks in total and denote $R_{i,j}$ as the retained accuracy on task $j$ after learning to task $i$, then
\begin{equation} \label{equ: transfer}
    \begin{split}
        &\text{ACC} = \frac{1}{T}\sum_{i=1}^T R_{T,i} \\
        &\text{BWD} = \frac{1}{T}\sum_{i=1}^T R_{T,i} - R_{i,i} \\
        &\text{FWD} = \frac{1}{T}\sum_{i=1}^T R_{i,i} \\
    \end{split}.
\end{equation}
It follows that $\text{ACC} = \text{BWD} + \text{FWD}$, which means BWD and FWD naturally trade-off between each other and therefore better algorithm will result in better ACC.
\begin{table*}[t]
    \centering
    \resizebox{\textwidth}{!}{%

    \begin{tabular}{r|ccc|ccc|ccc}
    \toprule
    Model & \multicolumn{3}{c}{MNIST Permutations} & \multicolumn{3}{c}{MNIST Rotations} & \multicolumn{3}{c}{Split CIFAR100}\\
          & ACC($\uparrow$) & FWD($\uparrow$)  & BWD($\uparrow$)  & ACC($\uparrow$) & FWD($\uparrow$)  & BWD($\uparrow$) & ACC($\uparrow$) & FWD($\uparrow$)  & BWD($\uparrow$)\\
    \hline
    Single      & 55.18{\footnotesize $\pm$0.52} & 85.62{\footnotesize $\pm$0.08} & -30.43{\footnotesize $\pm$0.59}  & 64.74{\footnotesize $\pm$2.40}  & 87.65{\footnotesize $\pm$0.07} & -22.91{\footnotesize $\pm$2.43} & 68.69{\footnotesize $\pm$0.32} & 74.10{\footnotesize $\pm$0.34} & -5.41{\footnotesize$\pm$0.27}\\
    Independent & 56.66{\footnotesize $\pm$3.79} & N/A            & N/A           &  56.66{\footnotesize $\pm$3.79}  & N/A                             & N/A                            &  61.61{\footnotesize $\pm$0.10} & N/A & N/A   \\
    EWC         & 57.04{\footnotesize $\pm$0.85} & 85.48{\footnotesize $\pm$0.08} & -28.43{\footnotesize $\pm$0.92}  & 65.74{\footnotesize $\pm$1.54}  & 85.75{\footnotesize $\pm$0.13} & -20.01{\footnotesize $\pm$1.53} & 70.10{\footnotesize $\pm$0.34} & 72.06{\footnotesize $\pm$0.12} & -1.96{\footnotesize$\pm$0.27}\\
    MER         & \textbf{83.71{\footnotesize $\pm$0.11}} & \textbf{87.01{\footnotesize $\pm$0.05}} &  \textbf{-3.29{\footnotesize $\pm$0.06}}  & \textbf{88.95{\footnotesize $\pm$0.11}}  & \textbf{88.77{\footnotesize $\pm$0.05}} &   0.17{\footnotesize $\pm$0.14} & 62.24{\footnotesize $\pm$0.17} & 66.84{\footnotesize $\pm$0.30} & -4.60{\footnotesize$\pm$0.42}\\
    GEM         & 80.54{\footnotesize $\pm$0.09} & 85.16{\footnotesize $\pm$0.08} &  -4.62{\footnotesize $\pm$0.14}  & 86.72{\footnotesize $\pm$0.12}  & 86.16{\footnotesize $\pm$0.06} &   0.57{\footnotesize $\pm$0.15} & 73.60{\footnotesize $\pm$0.27} & 75.02{\footnotesize $\pm$0.42} & -1.42\footnotesize{$\pm$0.57}\\
    \hline
    d-GEM       & 81.23{\footnotesize $\pm$0.09} & 84.86{\footnotesize $\pm$0.07} &  -3.64{\footnotesize $\pm$0.11}  & 87.02{\footnotesize $\pm$0.17}  & 85.86{\footnotesize $\pm$0.07} &   \textbf{1.16{\footnotesize $\pm$0.20}} & 74.12{\footnotesize $\pm$0.16} & 74.48{\footnotesize $\pm$0.21} & -0.36\footnotesize{$\pm$0.31}\\
    m-GEM       & 80.68{\footnotesize $\pm$0.09} & 85.21{\footnotesize $\pm$0.08} &  -4.53{\footnotesize $\pm$0.13}  & 86.74{\footnotesize $\pm$0.12}  & 86.18{\footnotesize $\pm$0.08} &   0.56{\footnotesize $\pm$0.14} & 75.23{\footnotesize $\pm$0.24} & \textbf{75.27{\footnotesize $\pm$0.46}} & -0.03\footnotesize{$\pm$0.56}\\
    md-GEM      & 81.18{\footnotesize $\pm$0.12} & 85.22{\footnotesize $\pm$0.09} &  -4.05{\footnotesize $\pm$0.13}  & 87.01{\footnotesize $\pm$0.17}  & 86.00{\footnotesize $\pm$0.06} &   1.01{\footnotesize $\pm$0.19} & \textbf{75.53{\footnotesize $\pm$0.16}} & 75.21{\footnotesize $\pm$0.40} &  0.32\footnotesize{$\pm$0.45}\\
    approx-GEM  & 81.79{\footnotesize $\pm$0.10} & 85.22{\footnotesize $\pm$0.09} &  -4.05{\footnotesize $\pm$0.13}  & 87.01{\footnotesize $\pm$0.12}  & 85.96{\footnotesize $\pm$0.07} &   1.05{\footnotesize $\pm$0.12} & 75.35{\footnotesize $\pm$0.35} & 75.03{\footnotesize $\pm$0.24} &  \textbf{0.32\footnotesize{$\pm$0.42}}\\
    \bottomrule
    \end{tabular}
    }
    \caption{Performance of mGEM and various baseline methods on the three standard continual learning benchmarks. m-GEMs have 4 modules for MNIST datasets and 9 modules for Split CIFAR100. d-GEM has 2 modules for all datasets.}
    \label{tab:cl-benchmark}
\end{table*}

\subsection{Standard Continual Learning Benchmarks}
\label{sec:cl-benchmark}
We first apply mGEM to the three standard continual learning benchmarks used in~\citet{lopez2017gradient}. \textbf{MNIST Permutations} consists of a sequence of 20 tasks, each of which is generated by a fixed permutation of the MNIST pixels. \textbf{MNIST Rotations} is another variant of 20 consecutive tasks based on MNIST~\cite{lecun1998gradient}. Each task is generated by rotating the MNIST digits by a fixed angle between 0 and 180 degrees. Therefore, both MNIST Permutations and MNIST Rotations have 20 10-class classification tasks. \textbf{Split CIFAR100} splits the 100 classes from CIFAR dataset~\cite{krizhevsky2009learning} into 20 subtasks, each is a 5-class classification task. For MNIST Permutations and MNIST Rotations, we test on a two-layer neural network with 100 hidden units. Moreover, we adopt the original setting from~\cite{lopez2017gradient} where 256 random samples from each task is stored as the episodic memory. For each task, we train the model for 500 iterations with a batch size of 10 before progressing to the next task. For Split CIFAR100, we test on a larger network consists of a stack of 5 convolutional layers. We sample 64 samples from each task as the episodic memory, and we train each task for 100 iterations with a batch size of 25. For both GEM and mGEM, we pick the best $\q$ from $\{0.0, 0.05, 0.1, 0.2, 0.3, 0.5, 0.8,1.0\}$ using a validation set. We provide more detailed information of the architecture and hyperparameters in the Appendix.

Table~\ref{tab:cl-benchmark} shows the result on the three benchmarks. Note that on MNIST Permutations and MNIST Rotations, MER achieves the best final accuracy. But after a comprehensive search of hyper-parameters, we find MER performs much worse on the Split CIFAR100 dataset. We think there are mainly two reasons. First, as the tasks in MNIST Permutations and MNIST Rotations are highly correlated, MER's meta training objective helps improve the zero-shot generalization so that MER can achieve higher FWD. But the tasks within the Split CIFAR100 dataset have less correlations so the meta learning objective becomes less effective. Secondly, meta learning is known to be data hungry. Therefore, on dataset like Split CIFAR100 which is more challenging but has fewer samples, it becomes harder for a meta learning algorithm like MER to learn. On the other hand, notice that on all datasets, mGEM outperforms GEM and improves BWD, which consolidates that adding more restrictions helps the model forget less. This effect is less prominent on the two MNIST datasets because of the size of the architecture. But on split CIFAR100, mGEM improves the average accuracy by $\bm{1.93}$.

\begin{figure}[t]
    \centering
    \includegraphics[width=\columnwidth]{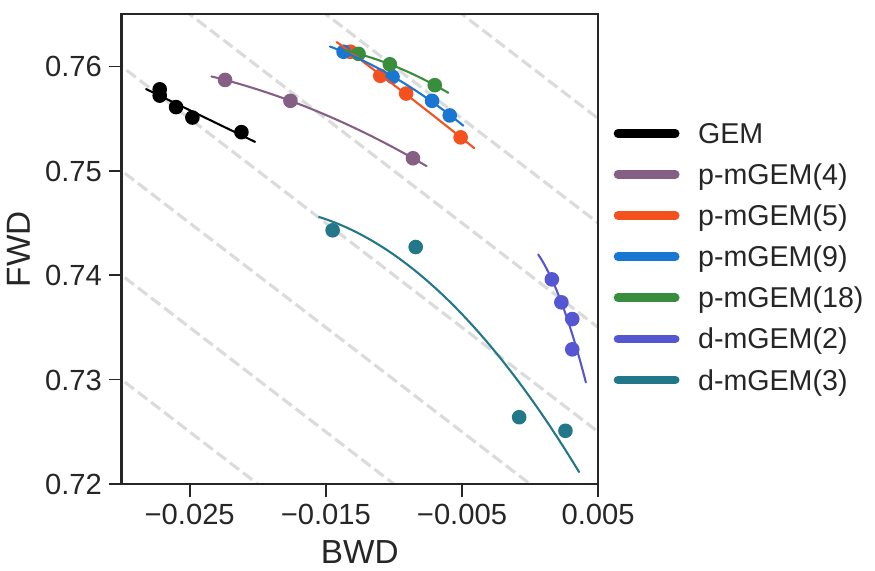}
    \vspace{-20pt}
    \caption{GEM and mGEM's Pareto Frontier of FWD and BWD on Split CIFAR100. The dashed gray lines are linear functions $y = -x + c$, where $c \in \mathbb{R}$. Thus points on the same dashed line have the same ACC. Numbers in the parenthesis denote how many modules we have for mGEM.}
    \label{fig:pareto}
    \vspace{-20pt}
\end{figure}

\textbf{Trade-off between FWD and BWD} To analyze how mGEM helps improve the overall performance, we conduct the ablation study that plots mGEMs with different number of modules. The result is shown in Figure~\ref{fig:pareto}. Here, the x and y axis correspond to BWD and FWD. p-mGEM($n$) denotes that we split the full network into $n$ modules. For instance, since the network considered in Split CIFAR100 has 5 convolutional layers plus 4 batchnorm layers, p-mGEM($9$) essentially treat every convolutional and batchnorm layer as a single module. Similarly, d-mGEM($n$) indicates we split the episodic memory into $n$ equal parts, which forms $n$ gradient constraints. As we can see from Figure~\ref{fig: pareto}, p-mGEM reaches better trade-off between FWD and BWD as $n$ increases, and we see no further improvement when $n$ is greater than $9$. d-mGEM(2) performs better than GEM but d-mGEM(3)'s performance degrades. This is likely due to the fact that d-mGEM(3) has 3 constraints per episodic memory, which over shrinks the search space.
\begin{figure*}[t]
    \centering
    \includegraphics[width=\textwidth]{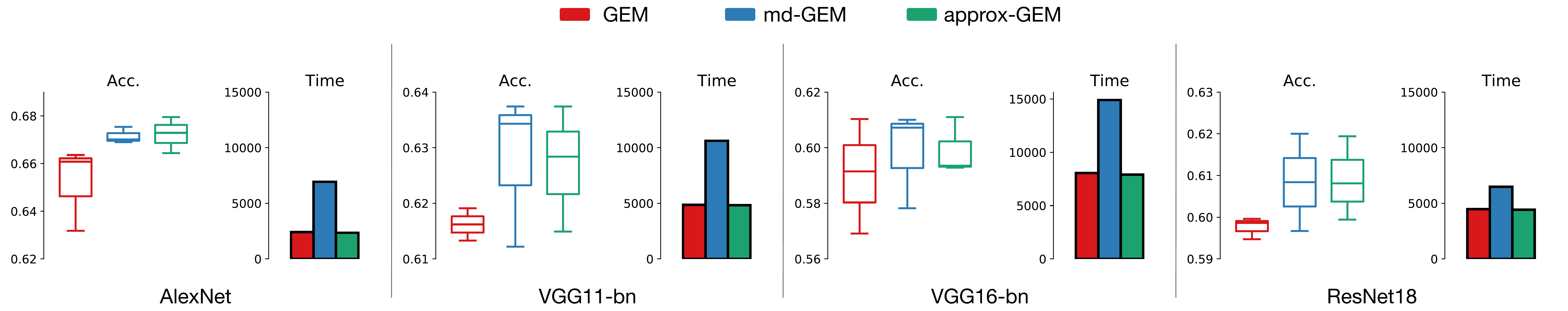}
    \vspace{-15pt}
    \caption{Performance of mGEM versus GEM on four commonly used neural architectures. The underlying continual learning task is the 10-way Split CIFAR100 benchmark. Here we report 3 runs with 3 different seeds for each architecture.}
    \label{fig:arch}
    \vspace{-10pt}
\end{figure*}

\textbf{mGEM on different architectures} Now that we observe mGEM improves over GEM, we wonder whether the performance is sensitive to the architecture we use. Therefore, we apply mGEM to 4 commonly used deep architectures: AlexNet~\cite{NIPS2012_c399862d}, VGG-11 with batchnorm layers (VGG11-bn), VGG-16 with batchnorm layers (VGG16-bn)~\cite{simonyan2014very}, and ResNet-18~\cite{he2016deep}. Since these architectures are large in size, we conduct experiments on the 10-way split of the Split CIFAR100, e.g., the CL problem consists of a sequence of 10 10-way classification tasks. Results are shown in Figure~\ref{fig:arch}. As we can see, md-GEM consistently outperforms GEM across all architectures. Meanwhile, with the approximation method in Section~\ref{sec:approx-GEM}, approx-GEM results in similar or less computation time than GEM but higher ACC.\footnote{We apply approx-GEM on each module sequentially. Further speed up can be achieved if module-wise computation is paralleled.}
\vspace{-5pt}
\subsection{Five Digits and DomainNet Dataset}
\begin{figure}[t]
    \centering
    \includegraphics[width=\columnwidth]{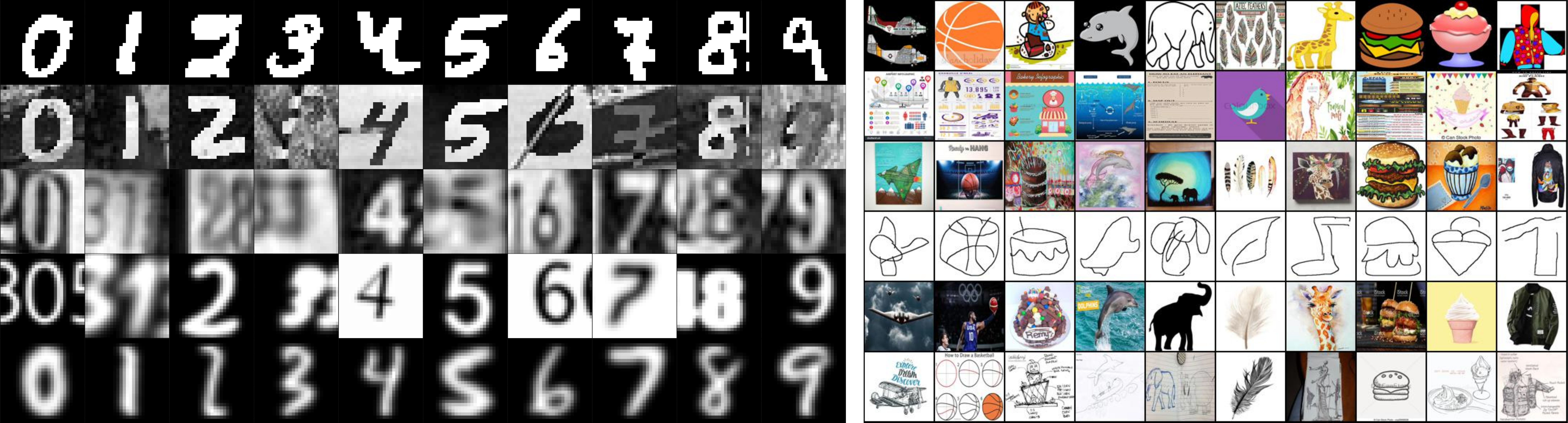}
    \vspace{-10pt}
    \caption{Sampled images from the Digit-Five (\textbf{left}) and DomainNet (\textbf{right}) datasets.}
    \label{fig:dataset}
    \vspace{-15pt}
\end{figure}

In all of the above continual learning benchmarks, the sequence of tasks originally are sampled from the same domain. In real-world application of the continual learning, the learner will more likely to encounter tasks from different domains. As a result, in this section we further apply mGEM to datasets that have more severe domain shift.
Particularly, we consider the Digit-Five dataset and the DomainNet dataset~\cite{peng2019moment}, both of which are standard benchmarks for the domain adaptation problem. ``Digit-Five" dataset indicates five popular digit datasets (MNIST~\cite{lecun1998gradient}, MNIST-M~\cite{ganin2015unsupervised}, Synthetic Digits~\cite{ganin2015unsupervised}, SVHN~\cite{netzer2011reading}, and USPS). DomainNet~\cite{peng2019moment} consists of 6 domains (clipart, painting, quickdraw, infograph, real, sketch) and each domain has 345 classes of common objects. Example data are provided in Figure~\ref{fig:dataset}.

For the Digit-Five dataset, we randomly generate 5 different permutations of the 5 datasets with each dataset containing 5K randomly sampled images. Then we apply the CL algorithms sequentially over the 5 datasets. For each dataset, we train the model for 20 epochs with a batch size of 256. For the DomainNet dataset, we train over any pairs of different domains, which results in 30 (6$\times$5) possible ways of transferring. For each way of transferring, we randomly sample 10 objects and form a 2-way 5-class classification tasks and apply the CL algorithms on it. We repeat this process 3 times for each way of transferring. So there are 90 continual learning problems in total. During training, we train on each domain with a batch size 32. For evaluation, we report the ACC, FWD and BWD, all averaged over the number of runs (5 in Digit-Five and 90 in DomainNet). For d-mGEM, we always split the episodic memory into 2 parts. For p-mGEM, we always treat each convolutional and batchnorm layer as a module. Results are summarized in Table~\ref{tab:5digit}.
\begin{table}[t]
\centering{}%
\begin{tabular}{r|ccc}
\toprule
\multirow{2}{*}{Model} & \multicolumn{3}{c}{Digit-Five Dataset}\tabularnewline
            &      ACC($\uparrow$)          & FWD($\uparrow$)  & BWD($\uparrow$)\tabularnewline
\hline 
Single      & 62.98{\footnotesize$\pm$6.66} & 67.47{\footnotesize$\pm$4.74} & -4.49{\footnotesize$\pm$9.85}\tabularnewline
Independent & 69.34{\footnotesize$\pm$1.49} & N/A & N/A\tabularnewline
MER         & 71.64{\footnotesize$\pm$4.20} & 72.48{\footnotesize$\pm$4.96} & -0.83{\footnotesize$\pm$6.03}\tabularnewline
EWC         & 66.79{\footnotesize$\pm$4.22} & 69.91{\footnotesize$\pm$4.70} & -3.12{\footnotesize$\pm$6.62}\tabularnewline
GEM         & 71.24{\footnotesize$\pm$3.57} & 72.53{\footnotesize$\pm$5.55} & -1.30{\footnotesize$\pm$5.29}\tabularnewline
\hline 
d-GEM       & 73.74{\footnotesize$\pm$2.79} & 73.61{\footnotesize$\pm$5.03} &  1.30{\footnotesize$\pm$4.30}\tabularnewline
m-GEM       & 72.62{\footnotesize$\pm$3.70} & 72.57{\footnotesize$\pm$5.51} &  0.06{\footnotesize$\pm$4.92}\tabularnewline
md-GEM      & \textbf{76.13{\footnotesize$\pm$2.43}} & 74.56{\footnotesize$\pm$4.96} &  \textbf{1.57{\footnotesize$\pm$4.33}}\tabularnewline
approx-GEM  & 74.31{\footnotesize$\pm$3.58} & \textbf{74.67{\footnotesize$\pm$4.76}} & -0.36{\footnotesize$\pm$6.54}\tabularnewline
\hline
     & \multicolumn{3}{c}{DomainNet} \\
    \hline 
    Single      & 38.80{\footnotesize$\pm$1.02} & 49.68{\footnotesize$\pm$1.06} & -10.88{\footnotesize$\pm$0.52}\\
    Independent & 48.25{\footnotesize$\pm$1.02} & N/A &   N/A\\
    MER         & 36.13{\footnotesize$\pm$1.06} & 41.70{\footnotesize$\pm$1.09} &  -5.57{\footnotesize$\pm$0.29}\\
    EWC         & 39.07{\footnotesize$\pm$1.02} & 49.76{\footnotesize$\pm$1.07} &  -10.70{\footnotesize$\pm$0.50}\\
    GEM         & 42.33{\footnotesize$\pm$1.14}  & 49.76{\footnotesize$\pm$1.06}           & -7.43{\footnotesize$\pm$0.37} \\
    \hline 
    d-GEM       & 43.75{\footnotesize$\pm$1.11}  & 49.95{\footnotesize$\pm$1.05}           & -6.20{\footnotesize$\pm$0.33} \\
    m-GEM       & 42.55{\footnotesize$\pm$1.12}  & 49.81{\footnotesize$\pm$1.04}           & -7.26{\footnotesize$\pm$0.36} \\
    md-GEM      & 44.12{\footnotesize$\pm$1.10}  & 50.01{\footnotesize$\pm$1.05}           & -5.89{\footnotesize$\pm$0.31} \\
    approx-GEM  & \textbf{44.20{\footnotesize$\pm$1.10}}  & \textbf{50.01}{\footnotesize$\pm$1.05}  & \textbf{-5.81{\footnotesize$\pm$0.32}} \\ 
    \bottomrule
\end{tabular}
\vspace{-5pt}
\caption{Performance of mGEM versus various baseline methods on Digit-Five (\textbf{top}) and DomainNet (\textbf{bottom}) datasets.}
\label{tab:5digit}
\vspace{-15pt}
\end{table}
From Table~\ref{tab:5digit}, we see that all mGEM methods consistently outperforms GEM and other baseline methods by improving the BWD significantly. Interestingly, we observe that mGEM do not hurt and can even improve FWD as well.

\vspace{-10pt}
\section{Conclusion}
In this paper, we analyze how the Gradient Episodic Memory (GEM) balances remembering old and learning new knowledge. Specifically, we show that restricting the search space of the update direction essentially reduces the generalization gap caused by limited size of episodic memory. Based on the analysis, we propose two novel ways of introducing additional gradient restrictions to achieve better Pareto Frontiers. Experiments on standard continual learning benchmarks and multi-domain datasets demonstrate that the proposed methods outperform GEM consistently.

\nocite{langley00}

\bibliography{reference}
\bibliographystyle{icml2021}
\onecolumn
\appendix

\section{Experiment Details}
\subsection{Details about Figure 2}
In the experiment section (Section 6), we adopt the conventional metrics used in previous continual learning literature to evaluate the performance of different methods. These metrics, e.g. ACC, BWD, and FWD, are all computed based on the learning accuracy, which might not directly correspond to the training objective, which is based on the inner product between gradients. As a result, in Figure 2, we directly plot the trade-off between forward and backward transfer using the computed inner products between the update direction and the gradient on the current and previous tasks. Specifically, in Figure 2, we calculate these inner products after the model learns over the second task, i.e. $s=1$ and $t=2$. Therefore, the x-axis corresponds to the average inner products between the update direction $\z$ and $\g_s$, and similarly the y-axis corresponds to the average inner products between the update direction $\z$ and $\g_t$.

\subsection{Dataset Details}
For the Digit-Five dataset, we resize each image from each dataset to 28$\times$28 resolution (same as MNIST input). Then we normalize all image pixels to have $0$ mean and $1$ standard deviation. For DomainNet dataset, we resize the images to 96$\times$96 resolution and normalize all images to the range $[0,1]$.

\subsection{Network Architecture}
For the experiments on the conventional CL benchmarks, we adopt the neural network with 2 hidden layers, each with 100 neurons for both MNIST Permutations and MNIST Rotations datasets. For Split CIFAR100, the network architecture is Conv2d(3, 128, 4, 2, 1), Conv2d(128, 256, 4, 2, 1), Conv2d(256, 512, 4, 2, 1), Conv2d(512, 1024, 4, 2, 1), Conv2d(1024, 100, 2, 1, 0). Here, we represent a 2d convolution layer by Conv2d(number of input channels, number of output channels, kernel size, stride, padding). Between any two convolutional layers we insert a batchnorm layer and use the ReLU activation. The architecture for Digit-Five dataset is Conv2d(1,32,3,1,0), Conv2d(32,64,3,1,0), Fc(64{*}144,
128), Fc(128, 10). Here Fc(x, y) denotes a fully connected layer with input dimension x and output dimension y. The architecture on DomainNet dataset is Conv2d(3, 64, 3, 2, 1), Conv2d(64, 128, 3, 2, 1), Conv2d(128, 256, 3,2, 1), Conv2d(256, 512, 3, 2,1), Conv2d(512, 512, 3,2,1), Conv2d(512, 10, 3, 1, 0).

\section{Proof}

\subsection{Proof of Proposition 1}

Notice that $\left\langle \g_{s},\z\right\rangle =\E_{(x,y)\sim\mathcal{D}_{s}}\left\langle \nabla_{\th}\ell(f_{\th},x,y),\z\right\rangle $
and $\left\langle \hat{\g}_{s},\z\right\rangle =\E_{(x,y)\sim\hat{\mathcal{D}}_{s}}\left\langle \nabla_{\th}\ell(f_{\th},x,y),\z\right\rangle $.
Define $R=\sup_{\z\in\Z}\left\Vert \z\right\Vert \left\Vert \g_{s}\right\Vert _{2,\infty}$
and consider the function class 
\[
H=\left\{ h_{\z}(x,y)=\frac{\left\langle \z,\nabla_{\th}\ell(f_{\th},x,y)\right\rangle }{R}:\z\in\Z\right\} .
\]
Notice that $\sup_{\z\in\Z,x,y}\left\langle \z,\nabla_{\th}\ell(f_{\th},x,y)\right\rangle \le\sup_{\z\in\Z}\left\Vert \z\right\Vert \left\Vert \g_{s}\right\Vert _{2,\infty}.$
Using uniform concentration inequality, given any probability, for
any $\delta>0$, with probability at least $1-\delta$, we have, for
any $h_{\z}\in H$, 
\[
\E_{(x,y)\sim\mathcal{D}_{s}}h_{\z}(x,y)\ge\E_{(x,y)\sim\mathcal{\hat{D}}_{s}}h_{\z}(x,y)+2\RR_{\left|\mathcal{\hat{D}}_{s}\right|}[H]+\sqrt{\frac{\log(1/\delta)}{\left|\mathcal{\hat{D}}_{s}\right|}},
\]
where $\RR_{m}[H]$ is the expected Rademacher complexity of $H$
defined by 
\[
\RR_{m}[H]=\E\frac{1}{m}\E_{\sigma_{i}}\left[\sup_{z\in\Z}\sum_{i=1}^{m}\sigma_{i}\left\langle \nabla_{\th}\ell(f_{\th},x_{i},y_{i}),\z\right\rangle /R\right].
\]
Multiple $R$ on both sides gives that 
\[
\left\langle \g_{s},\z\right\rangle \ge\left\langle \hat{\g}_{s},\z\right\rangle +2\RR_{\left|\mathcal{\hat{D}}_{s}\right|}[\Z]+R\sqrt{\frac{\log(1/\delta)}{\left|\mathcal{\hat{D}}_{s}\right|}},
\]
where 
\[
\RR_{\left|\mathcal{\hat{D}}_{s}\right|}[\Z]=\E\frac{1}{m}\E_{\sigma_{i}}\left[\sup_{z\in\Z}\sum_{i=1}^{m}\sigma_{i}\left\langle \nabla_{\th}\ell(f_{\th},x_{i},y_{i}),\z\right\rangle \right]
\]
which gives the desired result.

\subsection{Proof of Proposition 2}
Before we start the proof of Proposition 2, we introduce the following lemma.

\begin{lemma}
Given any partition of the parameter $\th=[\th^{1},...,\th^{D}]$.
Suppose $\z^{*}$ is the solution of the constraint optimization problem
of p-mGEM: 
\begin{align*}
 & \min_{\z}\left\Vert \g_{t}-\z\right\Vert ^{2}\\
\text{s.t.} & \left\langle \hat{\g}_{s}^{d},\z^{d}\right\rangle \ge\gamma_{s}^{d},\ \forall d\in[D].
\end{align*}
And suppose $\bar{\z}^{d}$ is the solution of the following constraint
problem 
\begin{align*}
 & \min_{\z^{d}}\left\Vert \g_{t}^{d}-\z^{d}\right\Vert ^{2}\\
\text{s.t.} & \left\langle \hat{\g}_{s}^{d},\z^{d}\right\rangle \ge\gamma_{s}^{d},
\end{align*}
where $\g_{t}^{d}=\nabla_{\th^{d}}\L_{\mathcal{D}_{t}}[f_{\th}]$.
We have $\z^{*}=[\bar{\z}^{1},...,\bar{\z}^{d}]$.
\end{lemma}

\paragraph{Proof}

Without loss of generality, we assume all the constraint problems
have one unique solution. Suppose $\z^{*}\neq[\bar{\z}^{1},...,\bar{\z}^{d}]$.
Without loss of generality, we suppose $\z^{*1}\neq\bar{\z}^{1}$
and thus $\left\Vert \g_{t}^{1}-\bar{\z}^{1}\right\Vert ^{2}<\left\Vert \g_{t}^{1}-z^{*1}\right\Vert ^{2}$.
We define $\z'=[\bar{\z}^{1},\z^{*2},...,\z^{*d}]$. Notice that 
\begin{align*}
\left\Vert \g_{t}-\z'\right\Vert ^{2} & =\sum_{d=2}^{d}\left\Vert \g_{t}^{d}-\z^{*d}\right\Vert ^{2}+\left\Vert \g_{t}^{1}-\bar{\z}^{1}\right\Vert ^{2}<\left\Vert \g_{t}-\z^{*}\right\Vert ^{2}\ \text{and}\\
 & \left\langle \hat{\g}_{s}^{d},\z^{*d}\right\rangle \ge\gamma_{s}^{d},\ \forall d\in\{2,3,...,D\}\\
 & \left\langle \hat{\g}_{s}^{1},\bar{\z}^{1}\right\rangle \ge\gamma_{s}^{1}.
\end{align*}
This implies that $\z'$ is also in the feasible set of p-mGEM and
improves over $\z^{*}$. This makes contradiction.

Now we start prove the Proposition 2. We first prove the first statement. Without loss of generality, we assume all the constraint problems have one unique solution. 
\paragraph{Proof of $\left\langle \hat{\protect\g}_{s},\protect\z_{\text{MS}}^{*}\right\rangle \ge\left\langle \hat{\protect\g}_{s},\protect\z_{\text{}}^{*}\right\rangle $}

We start with prove $\left\langle \hat{\g}_{s},\z_{\text{MS}}^{*}\right\rangle \ge\left\langle \hat{\g}_{s},\z_{\text{}}^{*}\right\rangle $
with $\gamma_{s}\ge0$. We consider two cases. Case 1: $\left\langle \hat{\g}_{s},\z^{*}\right\rangle \ge\gamma_{s}$.
In this case $\z_{\text{MS}}^{*}=\z^{*}$. Otherwise, 
\begin{align*}
\left\Vert \g_{t}-\z_{\text{MS}}^{*}\right\Vert ^{2} & <\left\Vert \g_{t}-\z^{*}\right\Vert ^{2},\ \ \text{and}\\
\left\langle \hat{\g}_{s},\z_{\text{MS}}^{*}\right\rangle  & \ge\gamma_{S}\ge0.
\end{align*}
This implies that $\z_{\text{MS}}^{*}$ is also in the feasible set
of the constraint problem in GEM and $\z_{\text{MS}}^{*}$ improves
over $\z^{*}$, which makes contradiction. Case 2: $\left\langle \hat{\g}_{s},\z^{*}\right\rangle <\gamma_{s}$
and thus $\left\langle \hat{\g}_{s},\z_{\text{MS}}^{*}\right\rangle \ge\gamma_{s}>\left\langle \hat{\g}_{s},\z_{\text{}}^{*}\right\rangle .$
In both cases, we have $\left\langle \hat{\g}_{s},\z_{\text{MS}}^{*}\right\rangle \ge\left\langle \hat{\g}_{s},\z_{\text{}}^{*}\right\rangle $. 

\paragraph{Proof of $\left\langle \hat{\protect\g}_{s},\protect\z_{\text{p-mGEM}}^{*}\right\rangle \ge\left\langle \hat{\protect\g}_{s},\protect\z_{\text{MS}}^{*}\right\rangle $}

We then prove $\left\langle \hat{\g}_{s},\z_{\text{p-mGEM}}^{*}\right\rangle \ge\left\langle \hat{\g}_{s},\z_{\text{MS}}^{*}\right\rangle $.
We consider two cases. Case 1: $\left\langle \hat{\g}_{s},\z_{\text{MS}}^{*}\right\rangle =\gamma_{s}$.
In this case, we have $\left\langle \hat{\g}_{s},\z_{\text{p-mGEM}}^{*}\right\rangle =\sum_{d=1}^{D}\left\langle \hat{\g}_{s},\z_{\text{p-mGEM}}^{*d}\right\rangle \ge\sum_{d=1}^{D}\gamma_{s}^{d}\ge\gamma_{s}.$
Case 2: $\left\langle \hat{\g}_{s},\z_{\text{MS}}^{*}\right\rangle >\gamma_{s}$.
Notice that this problem has strong duality and thus using KKT conditions,
we have 
\begin{align*}
\nabla_{\z}\left[\left\Vert \g_{t}-\z_{\text{MS}}^{*}\right\Vert ^{2}+\boldsymbol{v}^{*}\left(\left\langle \hat{\g}_{s},\z_{\text{MS}}^{*}\right\rangle -\gamma_{s}\right)\right] & =0\\
\boldsymbol{v}^{*}\left(\left\langle \hat{\g}_{s},\z_{\text{MS}}^{*}\right\rangle -\gamma_{s}\right) & =0.
\end{align*}
Notice that as $\left\langle \hat{\g}_{s},\z_{\text{MS}}^{*}\right\rangle >\gamma_{s}$,
we have $v^{*}=0$ and this implies that $\z_{\text{MS}}^{*}=g_{t}$.
For each $d\in[D]$, if $\left\langle \hat{\g}_{s}^{d},\z_{\text{MS}}^{*d}\right\rangle \le\gamma_{s}^{d}$
(here we partite $\z_{\text{MS}}^{*}$ by $\z_{\text{MS}}^{*}=[\z_{\text{MS}}^{*1},...,\z_{\text{MS}}^{*D}]$),
then we have $\left\langle \hat{\g}_{s}^{d},\z_{\text{p-mGEM}}^{*d}\right\rangle \ge\gamma_{s}^{d}\ge\left\langle \hat{\g}_{s},\z_{\text{MS}}^{*d}\right\rangle $.
If $\left\langle \hat{\g}_{s},\z_{\text{MS}}^{*d}\right\rangle >\gamma_{s}^{d}$,
then we know $\z_{\text{MS}}^{*d}=g_{t}^{d}$ is also the solution
of 
\[
\min_{\z^{d}}\left\Vert \g_{t}^{d}-\z^{d}\right\Vert ^{2}\text{s.t.}\left\langle \hat{\g}_{s}^{d},\z^{d}\right\rangle \ge\gamma_{s}^{d}.
\]
Using Lemma 1 we know that $\z_{\text{MS}}^{*d}=\z_{\text{p-mGEM}}^{*d}$.
For this case, we thus conclude that $\left\langle \hat{\g}_{s}^{d},\z_{\text{p-mGEM}}^{*d}\right\rangle \ge\left\langle \hat{\g}_{s},\z_{\text{MS}}^{*d}\right\rangle $.

\paragraph{Proof of $\left\langle \hat{\protect\g}_{s},\protect\z_{\text{d-mGEM}}^{*}\right\rangle \ge\left\langle \hat{\protect\g}_{s},\protect\z_{\text{MS}}^{*}\right\rangle $. }

We now prove $\left\langle \hat{\g}_{s},\z_{\text{d-mGEM}}^{*}\right\rangle \ge\left\langle \hat{\g}_{s},\z_{\text{MS}}^{*}\right\rangle $.
We consider two cases. Case 1: $\left\langle \tilde{\g}_{s}^{d},\z_{\text{MS}}^{*}\right\rangle \ge\gamma_{s}^{d}$
for all $d\in[D]$. In this case, using similar argument, we can show
that $\z_{\text{MS}}^{*}=\z_{\text{d-mGEM}}^{*}$. Case 2: $\left\langle \tilde{\g}_{s}^{d},\z_{\text{MS}}^{*}\right\rangle <\gamma_{s}^{d}$
for some $d\in[D]$. In this case 
\[
\left\langle \hat{\g}_{s},\z_{\text{MS}}^{*}\right\rangle \le\gamma_{s}\le\min_{d\in[D]}\gamma_{s}^{d}\le\min_{d\in[D]}\left\langle \tilde{\g}_{s}^{d},\z_{\text{d-mGEM}}^{*}\right\rangle \le\left\langle \hat{\g}_{s},\z_{\text{d-mGEM}}^{*}\right\rangle .
\]
Combine the two cases, we have the desired result.

\paragraph{Proof of the generalization gap statement}

For GEM with memory strength $\gamma_{s}$, we define the following
set 
\[
\Z_{\text{MS},n}(\gamma_{s})=\cup_{\hat{\mathcal{D}}_{s}}\left\{ z:\left\langle \nabla_{\th}\L_{\hat{\mathcal{D}}_{s}}[f_{\th}],\z\right\rangle \ge\gamma_{s}\right\} .
\]
Notice that GEM without memory strength is the case $\Z_{\text{MS},n}(0)$.
Here $\cup_{\hat{\mathcal{D}}_{s}}$ denotes the union of all possible
$\hat{\mathcal{D}}_{s}$ with $n$ i.i.d. training points. Similarly,
for p-mGEM and d-mGEM we define 
\begin{align*}
\Z_{\text{p-mGEM},n}(\gamma_{s}) & =\cup_{\hat{\mathcal{D}}_{s}}\left\{ \z:\left\langle \nabla_{\th^{d}}\L_{\hat{\mathcal{D}}_{s}}[f_{\th}],\z\right\rangle \ge\gamma_{s}^{d},\ \forall d\in[d]\right\} \\
\Z_{\text{d-mGEM},n}(\gamma_{s}) & =\cup_{\hat{\mathcal{D}}_{s}}\left\{ \z:\left\langle \nabla_{\th}\L_{\hat{\mathcal{D}}_{s}}[f_{\th}],\z\right\rangle \ge\gamma_{s}^{d},\ \forall d\in[d]\right\} .
\end{align*}
Notice that we have $\z^{*}\in\Z_{\text{MS},\left|\hat{\mathcal{D}}_{s}\right|}(0)$,
$\z_{\text{MS}}^{*}\in\Z_{\text{MS},\left|\hat{\mathcal{D}}_{s}\right|}(\gamma_{s})$,
$\z_{\text{p-mGEM}}^{*}\in\Z_{\text{p-mGEM},\left|\hat{\mathcal{D}}_{s}\right|}(\gamma_{s})$
and $\z_{\text{d-mGEM}}^{*}\in\Z_{\text{d-mGEM},\left|\hat{\mathcal{D}}_{s}\right|}(\gamma_{s})$.
Also notice that $\Z_{\text{MS},n}(\gamma_{s})\subseteq\Z_{\text{MS},n}(0)$,
$\Z_{\text{p-mGEM},n}(\gamma_{s})\subseteq\Z_{\text{MS},n}(\gamma_{s})$
and $\Z_{\text{d-mGEM},n}(\gamma_{s})\subseteq\Z_{\text{MS},n}(\gamma_{s})$.
This implies that 
\[
\RR_{\left|\mathcal{\hat{D}}_{s}\right|}[\Z_{\text{p-mGEM},\left|\mathcal{\hat{D}}_{s}\right|}(\gamma_{s})],\ \RR_{\left|\mathcal{\hat{D}}_{s}\right|}[\Z_{\text{d-mGEM},\left|\mathcal{\hat{D}}_{s}\right|}(\gamma_{s})]\le\RR_{\left|\mathcal{\hat{D}}_{s}\right|}[\Z_{\text{MS},\left|\mathcal{\hat{D}}_{s}\right|}(\gamma_{s})]\le\RR_{\left|\mathcal{\hat{D}}_{s}\right|}[\Z_{\text{MS},\left|\mathcal{\hat{D}}_{s}\right|}(0)].
\]
Also we have 
\[
\sup_{\z\in\Z_{\text{p-mGEM},\left|\mathcal{\hat{D}}_{s}\right|}(\gamma_{s})}\left\Vert \z\right\Vert ,\sup_{\z\in\Z_{\text{d-mGEM},\left|\mathcal{\hat{D}}_{s}\right|}(\gamma_{s})}\left\Vert \z\right\Vert \le\sup_{\z\in\Z_{\text{MS},\left|\mathcal{\hat{D}}_{s}\right|}(\gamma_{s})}\left\Vert \z\right\Vert \le\sup_{\z\in\Z_{\text{MS},\left|\mathcal{\hat{D}}_{s}\right|}(0)}\left\Vert \z\right\Vert .
\]
This gives the desired result.





\end{document}


\global\long\def\th{\boldsymbol{\theta}}%
\global\long\def\th{\boldsymbol{\theta}}%
\global\long\def\L{\mathcal{L}}%
\global\long\def\S{\boldsymbol{S}}%
\global\long\def\H{\text{H}}%
\global\long\def\R{\mathbb{R}}%
\global\long\def\Th{\boldsymbol{\Theta}}%
\global\long\def\RR{\mathfrak{R}}%
\global\long\def\E{\mathbb{E}}%
\global\long\def\Z{\mathcal{Z}}%
\global\long\def\T{\mathcal{T}}%
\global\long\def\D{\mathcal{D}}%
\global\long\def\X{\mathcal{X}}%
\global\long\def\Y{\mathcal{Y}}%
\global\long\def\P{\mathbb{P}}%
\global\long\def\z{\boldsymbol{z}}%
\global\long\def\g{\boldsymbol{g}}%
\global\long\def\G{\textbf{G}}%
\global\long\def\H{\textbf{H}}%
\global\long\def\v{\boldsymbol{v}}%
\global\long\def\q{\boldsymbol{q}}%

\twocolumn[
\icmltitle{
Appendix: Fine-Grained Gradient Restriction: A Simple Approach\\ for Mitigating Catastrophic Forgetting
}

%



\icmlsetsymbol{equal}{*}

\begin{icmlauthorlist}
\icmlauthor{Aeiau Zzzz}{equal,to}
\icmlauthor{Bauiu C.~Yyyy}{equal,to,goo}
\icmlauthor{Cieua Vvvvv}{goo}
\icmlauthor{Iaesut Saoeu}{ed}
\icmlauthor{Fiuea Rrrr}{to}
\icmlauthor{Tateu H.~Yasehe}{ed,to,goo}
\icmlauthor{Aaoeu Iasoh}{goo}
\icmlauthor{Buiui Eueu}{ed}
\icmlauthor{Aeuia Zzzz}{ed}
\icmlauthor{Bieea C.~Yyyy}{to,goo}
\icmlauthor{Teoau Xxxx}{ed}
\icmlauthor{Eee Pppp}{ed}
\end{icmlauthorlist}

\icmlaffiliation{to}{Department of Computation, University of Torontoland, Torontoland, Canada}
\icmlaffiliation{goo}{Googol ShallowMind, New London, Michigan, USA}
\icmlaffiliation{ed}{School of Computation, University of Edenborrow, Edenborrow, United Kingdom}

\icmlcorrespondingauthor{Cieua Vvvvv}{c.vvvvv@googol.com}
\icmlcorrespondingauthor{Eee Pppp}{ep@eden.co.uk}

\icmlkeywords{Machine Learning, ICML}

\vskip 0.3in
]



\printAffiliationsAndNotice{\icmlEqualContribution} 

\appendix
\section{Details about Figure 2}
In the experiment section (Section 6), we adopt the conventional metrics used in previous continual learning literature to evaluate the performance of different methods. These metrics, e.g. ACC, BWD, and FWD, are all computed based on the learning accuracy, which might not directly correspond to the training objective, which is based on the inner product between gradients. As a result, in Figure 2, we directly plot the trade-off between forward and backward transfer using the computed inner products between the update direction and the gradient on the current and previous tasks. Specifically, in Figure 2, we calculate these inner products after the model learns over the second task, i.e. $s=1$ and $t=2$. Therefore, the x-axis corresponds to the average inner products between the update direction $\z$ and $\g_s$, and similarly the y-axis corresponds to the average inner products between the update direction $\z$ and $\g_t$.

\section{Network Architecture}-
For the experiments on the conventional CL benchmarks, we adopt the neural network with 2 hidden layers, each with 100 neurons for both MNIST Permutations and MNIST Rotations datasets. For Split CIFAR100, we use a convolution neural network with 5 hidden layers with 128, 256, 512, 1024, and 100 neurons. Each convolution layer except the last one has no bias, a stride with 2, and a padding with 1, followed by a batchnorm layer. The last convolutional layer has the bias vector and has no padding. For experiments on the Digit-Five and the DomainNet datasets, we adopt a similar network architecture as the one we used in Split CIFAR100. Specifically, the architecture on Digit-Five dataset

\nocite{langley00}

\bibliography{reference}
\bibliographystyle{icml2021}



